\documentclass{article}
\usepackage{ijcai21}

\usepackage{times}
\usepackage{soul}
\usepackage{url}
\usepackage[utf8]{inputenc}
\usepackage[small]{caption}
\usepackage{graphicx}
\usepackage{amsmath}
\usepackage{amsthm}
\usepackage{booktabs}
\usepackage{algorithm}
\usepackage{algorithmic}

\usepackage[utf8]{inputenc} 
\usepackage[T1]{fontenc}    
\usepackage{url}            
\usepackage{amsfonts}       
\usepackage{nicefrac}       
\usepackage{microtype}      
\usepackage{bm}
\usepackage{array}
\usepackage{multirow}
\usepackage{subcaption}

\newtheorem{definition}{Definition}

\author{
Liang Chen$^1$
\and
Jintang Li$^1$\and
Qibiao Peng$^1$\and
Yang Liu$^1$\and
Zibin Zheng$^1$\footnote{Corresponding Author}\and
Carl Yang$^2$
\affiliations
$^1$Sun Yat-sen University\\ $^2$Emory University
\emails
chenliang6@mail.sysu.edu.cn,
\{lijt55, pengqb3, liuy296\}@mail2.sysu.edu.cn,\\
zhzibin@mail.sysu.edu.cn,
j.carlyang@emory.edu
}

\title{Understanding Structural Vulnerability in \\Graph Convolutional Networks}

\begin{document}

\maketitle

\begin{abstract}
Recent studies have shown that Graph Convolutional Networks (GCNs) are vulnerable to adversarial attacks on the graph structure. Although multiple works have been proposed to improve their robustness against such structural adversarial attacks, the reasons for the success of the attacks remain unclear. In this work, we theoretically and empirically demonstrate that structural adversarial examples can be attributed to the non-robust aggregation scheme (i.e., the weighted mean) of GCNs. Specifically, our analysis takes advantage of the breakdown point which can quantitatively measure the robustness of aggregation schemes. The key insight is that weighted mean, as the basic design of GCNs, has a low breakdown point and its output can be dramatically changed by injecting a single edge. We show that adopting the aggregation scheme with a high breakdown point (e.g., median or trimmed mean) could significantly enhance the robustness of GCNs against structural attacks. Extensive experiments on four real-world datasets demonstrate that such a simple but effective method achieves the best robustness performance compared to state-of-the-art models.
\end{abstract}

\section{Introduction}
Graph Convolutional Networks (GCNs)~\cite{DBLP:conf/iclr/KipfW17} have become increasingly popular for graph learning tasks, such as citation~\cite{DBLP:conf/iclr/VelickovicCCRLB18}, social~\cite{DBLP:conf/www/Fan0LHZTY19}, biological~\cite{DBLP:conf/iclr/XuHLJ19} and financial~\cite{DBLP:conf/icdm/WangQL0JWFYZY19} networks. GCNs generally follow a recursive neighborhood aggregation scheme, where at each iteration of aggregation, the representation of a node is generated by aggregating the representation of its neighbors, followed by a linear transformation and non-linear activation function. After $k$ iterations of aggregation, a node's representation can capture the structural information of its $k$-hop neighbors.

Despite the great success, there has been a surge of studies~\cite{DBLP:conf/kdd/ZugnerAG18,DBLP:conf/icml/DaiLTHWZS18,DBLP:conf/iclr/ZugnerG19,DBLP:conf/ijcai/Wu0TDLZ19,DBLP:conf/ijcai/XuC0CWHL19} show that GCNs are vulnerable to adversarial structural attacks. That is, slightly modifying the existence of either edges or nodes may lead to the wrong prediction. Even if the attacker does not have full knowledge of model parameters, they are still able to break the model~\cite{DBLP:conf/kdd/ZugnerAG18}. Such unreliable results provide the opportunity for attackers to exploit these vulnerabilities and restrict the applications of GCNs.

To resist such attacks, multiple methods have been explored in the literature including adversarial training~\cite{DBLP:conf/ijcai/XuC0CWHL19}, transfer learning~\cite{DBLP:conf/wsdm/TangLSYMW20}, employing Gaussian distributions to represent nodes~\cite{DBLP:conf/kdd/ZhuZ0019}, recovering potential adversarial perturbations~\cite{DBLP:conf/ijcai/Wu0TDLZ19},  and certifiable robustness~\cite{DBLP:conf/ecai/LiuPCZ20}. However, the design of defense algorithms is mostly based on empirical intuition, heuristics, and experimental trial-and-error. The reasons for the success of adversarial structural attacks and the vulnerability of GCNs remain unclear.


In this work, we aim to address a basic question in GCN's robustness: why graph convolutional networks are vulnerable to adversarial structural attacks? Specifically, we focus on the targeted evasion (test-time) attack: given a trained GCN (the model parameters are fixed) and a targeted node, the goal is to construct an adversarial example (by modifying edges) that fools the model. A new perspective is presented to explain such a phenomenon. We claim that:
\begin{center}
    \textit{The non-robust aggregation function leads to structural vulnerability of GCNs.}
\end{center}
Recall that GCNs recursively perform representation aggregation and transformation during the training phase. Consequently, if the aggregation function is sensitive to slight perturbations, the whole model built on such a function tends to be non-robust~\cite{xie2020gnns}. We formally characterize how vulnerable the classical aggregation function (i.e., weighted mean) of GCNs is through the theory of breakdown point~\cite{donoho1992breakdown,Huber:1254106}, which is often employed to measure the robustness of estimators by quantifying the influence of outliers in robust statistics. Our key insight is that the breakdown point of the weighted mean is low, indicating GCNs' aggregated representations can be easily modified by slight perturbations. Our hypothesis also provides a method for more robust GCNs: employing aggregation schemes with high breakdown points such as the median and trimmed mean. 

Our main contributions are summarized as follows:
\begin{itemize}
    \item We establish the connection between the robustness of aggregation functions and GCNs via the breakdown point theory. We reveal that the aggregation function of GCNs has a large influence on the robustness of GCNs against structural attacks.
    \item We propose trimmed mean and median aggregation functions with high breakdown points to improve GCNs' robustness against structural attacks.
    \item We validate our theory via experiments on four node classification datasets, where we compare the robustness and performance with the state-of-the-art models. Our results confirm that employing robust aggregation functions could achieve high model robustness without sacrificing predictive accuracy.
\end{itemize}

\section{Related Work}
Extensive studies have demonstrated that GCNs are vulnerable to adversarial structural attacks which can be classified into evasion (test-time)~\cite{DBLP:conf/ijcai/Wu0TDLZ19,DBLP:conf/icml/DaiLTHWZS18,DBLP:conf/kdd/ZugnerAG18} and poisoning (training-time)~\cite{DBLP:conf/kdd/ZugnerAG18,DBLP:conf/iclr/ZugnerG19,DBLP:conf/icml/BojchevskiG19}, non-targeted (whole system)~\cite{DBLP:conf/ijcai/XuC0CWHL19,DBLP:conf/iclr/ZugnerG19} and targeted (specific instances)~\cite{DBLP:conf/icml/DaiLTHWZS18,DBLP:conf/kdd/ZugnerAG18} attacks, according to the attack stage and goal \cite{chen2020survey}, respectively. Increasing efforts have been devoted to improve the robustness of GCNs.
Based on the strategy to enhance the robustness of GCNs, we can classify existing defense methods into the following three types: robust representation, robust detection, and robust optimization.

\paragraph{Robust representation.}
The central idea of these methods~\cite{DBLP:conf/nips/BojchevskiG19,DBLP:conf/kdd/ZugnerG19,DBLP:conf/ijcai/XuC0CWHL19} is to map nodes to a robust space such that even there exist perturbations, the predicted label of the target node remains the same. Note that these methods consider the robustness of GCNs under the transductive setting. Though the aggregation function is non-robust, they find a solution of robust space for considered finite nodes which are usually achieved by optimizing the worst-case loss. Since the exact worst-case perturbation is hard to compute, the quality of robust representations relies on the approximation of the worst-case perturbation such as robustness certificates~\cite{DBLP:conf/nips/BojchevskiG19,DBLP:conf/kdd/ZugnerG19} or adversarial examples~\cite{DBLP:conf/ijcai/XuC0CWHL19}.

\paragraph{Robust detection.}
Similar to the robust representation, the detection methods do not change the aggregation function of GCNs as well. Their goal is to identify the correct nodes to aggregate. Existing methods include Jaccard detection~\cite{DBLP:conf/ijcai/Wu0TDLZ19} which eliminating edges that connect nodes with the Jaccard similarity of features smaller than a given threshold, and singular value decomposition~\cite{DBLP:conf/wsdm/EntezariADP20} which preprocesses the graph with its low-rank approximation. 

\paragraph{Robust optimization.}
These methods employ robust optimization strategy, which is not sensitive to extreme embeddings, to train the graph model. PA-GNN~\cite{DBLP:conf/wsdm/TangLSYMW20} learns attention scores via transferring the knowledge from clean graphs. RGCN~\cite{DBLP:conf/kdd/ZhuZ0019} adopts Gaussian distributions as hidden representations and variance-based attention mechanism to perform graph convolution. SimPGCN~\cite{simpgcn} integrates the node features with the structure information and adaptively balance their influence on the aggregation process. One step further, Soft Medoid~\cite{GeislerZG20}, a robust message-aggregation function is proposed to improve the robustness of GCNs against structural perturbations. Our method is close to this work in that we employ a high breakdown point function to address the drawback of message passing scheme in GCNs.

In contrast to the above works, our work aims to find the explanation for the success of adversarial structural attacks. Specifically, we mainly focus on the targeted evasion attacks which are more practical in the real world. We show that the robustness of the aggregation function has a large influence on the robustness of GCNs. Based on this observation, we propose to use two simple but effective aggregations (i.e., the median and trimmed mean) which achieve state-of-the-art robustness compared to existing works.

\section{The Structural Vulnerability of GCNs}
\subsection{Graph Convolutional Networks}
First we introduce graph convolutional networks and notations under the context of the node classification task, where the goal is to predict the associated class label for each node. Let $G = (V, E)$ denote the graph with $X_v$ representing the node features of $v\in V$. GCN incorporates graph structure and node feature to learn representations. Specifically, it iteratively updates the representation of a node by aggregating messages of its neighbors. The $k$-th layer of a GCN can be formally defined as:
\begin{equation}\label{eq:gnn}
a_{v}^{(k)} = f\left(\left\{h_{u}^{(k-1)}:u\in \mathcal{N}_{v} \right\}\right),\
h_{v}^{(k)} = \sigma(a_{v}^{(k)}W^{(k)}),
\end{equation}
where $a_{v}^{(k)}$ and $h_{v}^{(k)}$ are the aggregated embedding and transformed embedding of the $k$-th layer. Initially, $h_{v}^{(0)} = X_{v}$. $\mathcal{N}_{v}$ denotes the set of nodes adjacent to $v$. Note that the graph includes self connections, thus $\mathcal{N}_{v}$ contains the target node itself. 
The aggregation function $f$ is the core design of GCNs which is defined as follows:
\begin{equation}\label{eq:aggregation}
a_{v}^{(k)} = \sum_{u\in \mathcal{N}_{v}}w_{uv}\ h_{u}^{(k-1)}, \ w_{uv} = \frac{1}{\sqrt{|\mathcal{N}_{u}||\mathcal{N}_{v}|}},
\end{equation}
where $w_{uv}$ is the static weight between node $u$ and $v$.

\subsection{Empirical Analysis}\label{sec:em}
To investigate the vulnerability of GCNs against structural perturbations, we conduct an empirical study on Cora and Citeseer (dataset statistics can be found in Section~\ref{sec:data}). Note that the goal of existing attack methods~\cite{DBLP:conf/kdd/ZugnerAG18,DBLP:conf/icml/DaiLTHWZS18} is to approximate the worst-case perturbations given a perturbation budget (i.e., the number of modified edges). In our empirical study, we first train a two-layer GCN, then we enumerate all possible structural perturbations with the perturbation budget as one (i.e., modifying only one single edge that directly connected to the target node) and check whether such perturbations will change the predictive class for the target node. This attack finds the exact worst-case perturbation of direct perturbation. All nodes in datasets are target nodes. Our modifications are classified into three types -- Injection (i.e., injecting a new edge), Deletion (i.e., deleting an existing edge), and Both (i.e., using both injection and deletion).

\begin{figure}[t]
\centering
\hspace{0.1cm}
\includegraphics[width=0.8\linewidth]{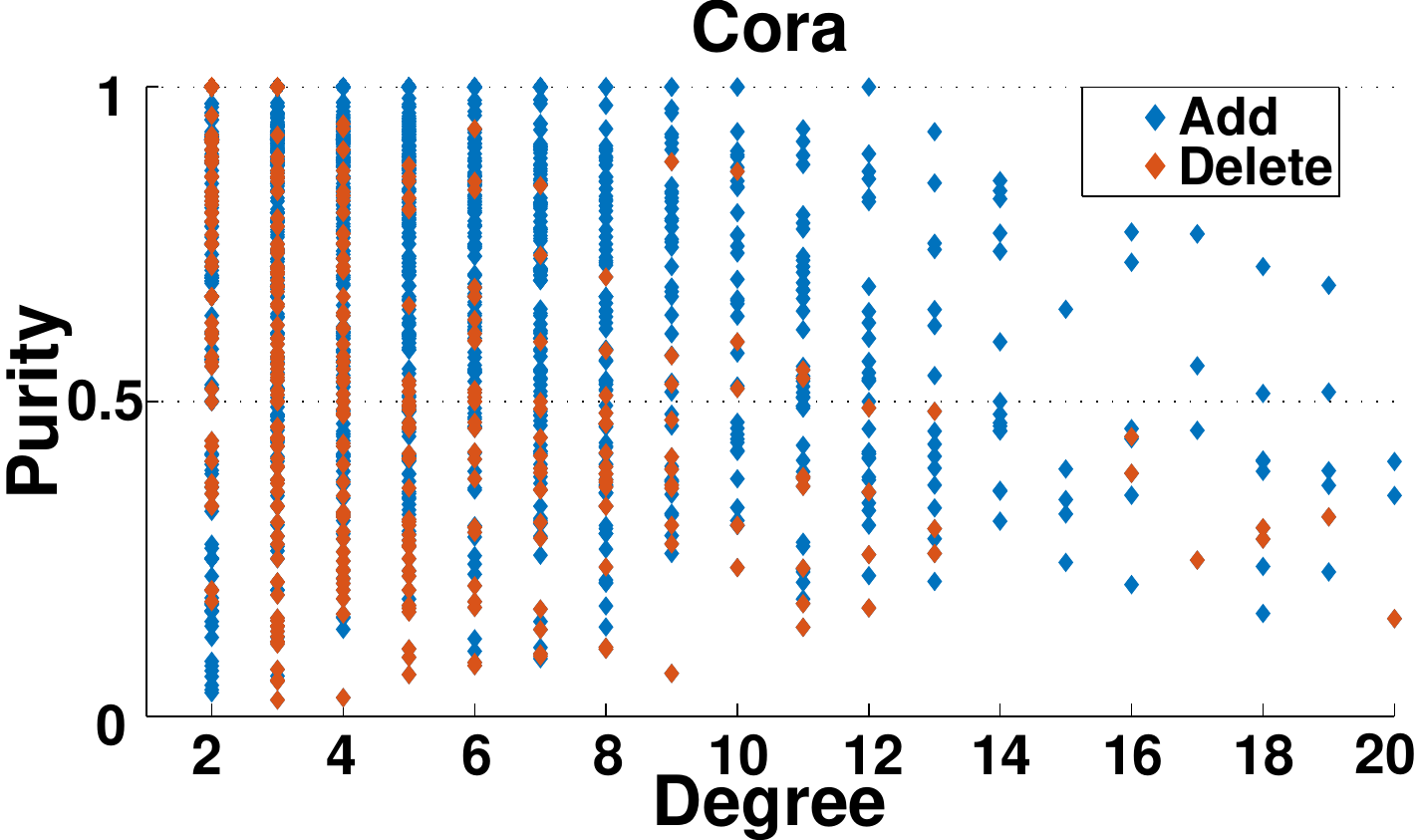}
\caption{The degree and purity of successfully attacked nodes with single edge perturbation.} \label{fig:em_f} 
\end{figure}

Figure~\ref{fig:em_f} plots the \textbf{degree} and \textbf{purity} (i.e. the percentage of nodes with the same class in the target node’s two-hop neighborhood) of nodes and Table~\ref{table:em_t} shows the percentage of successfully attacked nodes. From experimental results, we have two observations: (i) Most nodes can be successfully attacked by modifying only one edge, even though most nodes have high purity and degree\footnote{Nodes with high degree and purity are more robust according to \cite{DBLP:conf/kdd/ZugnerAG18,DBLP:conf/nips/BojchevskiG19}. }. For example, in Cora, the node with degree=12 and purity=1 is still successfully attacked. (ii) Injecting an edge can attack more nodes than deleting one. Moreover, we can observe that the percentage of the method Injection and Both is equal, indicating if a node can not be broken by injection, it can not be broken by deletion as well. \textit{These results reveal a finding that injection is an operation with stronger attack power, compared with deletion}. The empirical results are consistent with the previous study \cite{DBLP:journals/corr/abs-2003-00653,DBLP:conf/nips/BojchevskiG19} that attackers tend to add edges rather than remove them and reduce the purity of nodes in order to achieve the adversarial goals.  Based on these observations, we offer a further explanation to the structural vulnerability of GCNs via breakdown point theory (Section \ref{GCN_vulnerability}), and improve the robustness of GCNs with more robust aggregation functions (Section \ref{our_methods}).

\begin{table}[t]
\centering
\begin{tabular}{c|c|c|c} 
\hline  
Dataset & Injection & Deletion & Both\\
\hline
Cora & 79.57 & 14.26 & 79.57\\
Citeseer & 84.66 & 13.77 & 84.66\\
 \hline 
\end{tabular} \vspace{12pt}
\vspace{-0.5cm}
\caption{The percentage (\%) of successfully attacked nodes.}\label{table:em_t}
\vspace{-0.1cm}
\end{table}

\section{Robustness Analysis}
In this section, we first present our robustness analytical framework, and then we use this framework to explain the success of direct and indirect structural attacks.

\subsection{Analytical Framework}
\begin{figure}[t]
\centering
\includegraphics[width=\linewidth]{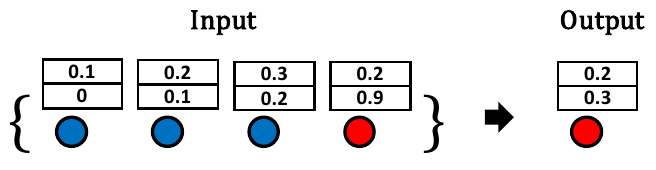}
\caption{A simple example of mean aggregation. The output is influenced by a single perturbation (red node).} \label{fig:aggregation} 
\vspace{-0.3cm}
\end{figure}

\paragraph{From GCNs to aggregation function.} 
Recall that a core operation of GCNs is to recursively update each node's embedding by aggregating the ones of their neighbors. Since structural adversarial attacks directly change the input of aggregation function, the robustness of the aggregation function is closely related to the robustness of GCNs. Specifically, injecting/deleting an edge of a node will add/remove an embedding from the input set (i.e., the set of embeddings needed to be aggregated). If a GCN is built on a non-robust aggregation function, the whole model tends to be non-robust as well. Figure~\ref{fig:aggregation} is a simple example of mean aggregation. As can be seen from Figure~\ref{fig:aggregation}, although most input nodes are blue class and there is only one single perturbation, the output is classified to the red class. Thus it is crucial to identify whether the existing aggregation function of GCNs is robust and whether employing a more robust aggregation can improve the robustness against structural attacks.


\paragraph{Robustness measure.} 
To address the above problems, the major challenge is how to quantitatively measure the robustness of aggregation functions. Previous work~\cite{DBLP:conf/ijcai/Wu0TDLZ19} has shown that the perturbed edges tend to connect nodes with low feature similarity. In other words, the adversarially injected node can be viewed as an outlier compared to other true neighbors. Now the key problem is that outliers are hard to define. Following the existing work~\cite{donoho1992breakdown,Huber:1254106}, the robustness of a function can be estimated under the extreme case. That is, whether adding multiple infinities values to the input data will cause the output to infinity. Formally, we employ the breakdown point to quantitatively measure the robustness of a given aggregation function, which refers to the maximum fraction of perturbations that an aggregation scheme can tolerate before incurring arbitrarily high error. Let $\mathcal{N}_{v}$ denote the set of input nodes. If there exists a perturbation set $\widetilde{\mathcal{N}}_{v}$ such that $f(\mathcal{N}_{v}\cup\widetilde{\mathcal{N}}_{v}) - f(\mathcal{N}_{v})$ is arbitrarily large, the aggregation scheme is said to break down under the contamination fraction $|\mathcal{N}_{v}|/(|\mathcal{N}_{v}|+|\widetilde{\mathcal{N}}_{v}|)$. Fortunately, since the aggregation functions considered in this work are element-wise (e.g., the aggregation function of GCNs is element-wise weighted mean), we can reduce our analysis from high-dimension to 1-dimension. Therefore, we assume the output of $f$ is a scalar corresponding to a certain dimension.
The definition is as follow:

\begin{figure*}[t]
\centering
\includegraphics[width=0.7\textwidth]{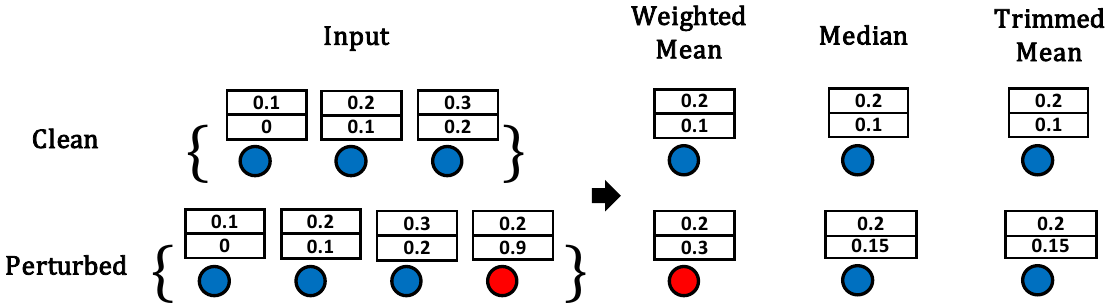}
\caption{A simple example of the weighted mean, median, and trimmed mean aggregation. The trimmed mean discards the largest and smallest value.} \label{fig:attack} 
\vspace{-0.5cm}
\end{figure*}

\vspace{0.2cm}
\begin{definition}
Breakdown point $\epsilon(f, \mathcal{N})$ is defined as the smallest contamination fraction under which the aggregation function $f$ breaks down:
\begin{equation}\label{eq:breakdown}
\nonumber
\min_{m\in\mathbb{N}}\left\{\frac{m}{|\mathcal{N}_{v}|+m}:\sup\limits_{\widetilde{\mathcal{N}}_{v}:|\widetilde{\mathcal{N}}_{v}|=m}  |f(\mathcal{N}_{v}\cup\widetilde{\mathcal{N}}_{v}) - f(\mathcal{N}_{v})|=\infty  \right\}
\end{equation}
\end{definition}
Breakdown point has been widely used in robust statistics to study the estimators such as mean, median, and trimmed mean. \textit{A larger breakdown point indicates better robustness}. Since the size $m$ of non-empty set $\widetilde{\mathcal{N}}_{v}$ is always greater than zero (i.e. $m\geq 1$), the smallest value of breakdown point is $1/(|\mathcal{N}_{v}| + 1)$. The value of the breakdown point can not exceed $1/2$ (i.e., $m=|\mathcal{N}_{v}|$) because if the number of perturbations is larger than the normal nodes, it is not possible to distinguish between perturbations and true nodes. Note that we only define the situation of injecting new nodes. The reason is that deleting $k$ ($k<|\mathcal{N}_{v}|$) nodes of aggregation function (e.g., the weighted mean) will not lead to an infinity value, which indicates that injecting nodes can lead to a larger change of output than deleting nodes. This also explains the experimental results of Table~\ref{table:em_t} and is consistent with observations in the previous work~\cite{DBLP:conf/nips/BojchevskiG19}.

\subsection{Understanding Structural Vulnerability of GCNs}
\label{GCN_vulnerability}
\paragraph{The instability of weighted mean.} We first study the robustness of the weighted mean (i.e., the aggregation function of GCNs). Specifically, when there exists a perturbation set, the output of weighted mean is as follows:
\begin{equation}
a_{v} = \sum_{u\in \mathcal{N}_{v}}w_{uv}\ h_{u} + \sum_{\tilde{u}\in \widetilde{\mathcal{N}}_{v}}w_{\tilde{u}v}\ h_{\tilde{u}}
\end{equation}
For simplicity, we omit the superscript $(k)$. $h_{u}\in\mathbb{R}$ and $h_{\tilde{u}}\in\mathbb{R}$ denote the normal and infinity value, respectively.  Since the weight $w_{\tilde{u}v}$ is always positive in GCNs, the output $a_{v}$ is linearly correlated to the perturbations $h_{\tilde{u}}$. Therefore, even a single infinity value of perturbation leads to an infinity output. The breakdown point of the weighted mean is $1/(|\mathcal{N}_{v}| + 1)$ which is the smallest value of breakdown point, indicating the weighted mean is highly non-robust against structural perturbations.

Given a target node, structural adversarial attacks usually contain two types of perturbations: (i) direct perturbations that injecting direct edges to the target node; (ii) indirect perturbations that modifying edges that are not directly connected to the target node. Our experimental results and previous works~\cite{DBLP:conf/kdd/ZugnerAG18,DBLP:conf/iclr/ZugnerG19} have shown that GCNs are vulnerable to both direct and indirect attacks. Based on the non-robustness of weighted mean, we can explain the success of direct and indirect perturbations as follows.

\paragraph{Direct perturbations.} 
Direct perturbations directly add perturbations to the input set of the weighted mean. Since the breakdown point of the weighted mean is small, it is possible to change the output via a small number of perturbations. This conclusion is consistent with observations of our empirical study in Section~\ref{sec:em} where the prediction of most nodes can be broken by a single perturbation. In practice, we can not find a node that has an infinity value. However, a node with a rather different value is enough to destroy the GCNs. As Section~\ref{sec:em} shows, for most nodes (around 80\% in Cora and Citeseer), a single perturbation is able to change the classification of a model that is trained normally.



\paragraph{Indirect perturbations.} 
Indirect perturbations do not change the neighboring edges of the target node, thus the number of neighbor embeddings that need to be aggregated is the same before and after attacks. Since the weighted mean is non-robust and GCNs recursively aggregate the local information of nodes, the target node will be attacked due to cascading failures. That is, the representations of perturbed neighbors are first broken, then the affected neighbors will influence the result of the target node.
Overall, the aggregation function of GCNs is non-robust. Since existing attacks~\cite{DBLP:conf/kdd/ZugnerAG18,DBLP:conf/icml/DaiLTHWZS18} only limit the number of changed edges and have no constraints on features of changed nodes, it is possible to successfully perform direct or indirect attacks as long as the features of perturbations vary greatly.

\section{Building Robust Graph Convolutional Networks}
\label{our_methods}
To resist structural attacks, an intuitive idea is to employ aggregation functions that have a high breakdown point. In this section, two alternative aggregation functions are introduced -- median and trimmed mean aggregation. Although such aggregations are simple, they achieve state-of-the-art performance according to our experimental results in Section~\ref{sec:analysis}. Figure~\ref{fig:attack} is a comparison of the weighted mean, median, and trimmed mean aggregation. As can be seen from Figure~\ref{fig:attack}, all strategies can learn the right representation from clean data. But when there exists a perturbation, the output of the weighted mean is easier to be affected than the median and trimmed mean aggregation. 

\paragraph{Median aggregation.}
According to the above analysis, we can improve the robustness of GCNs against adversarial structural perturbations via a more robust aggregation scheme. A simple but effective solution is to replace the weighted mean to median aggregation. Suppose $\left[h_{u}\right]_{u=1}^{n}$ is sorted and $n$ is the size of $\mathcal{N}_{v}\cup\widetilde{\mathcal{N}}_{v}$, we employ the following element-wise median to aggregation embeddings in Equation (\ref{eq:median}).
\begin{equation}\label{eq:median}
a_{v} = \begin{cases}
(h_{n/2} + h_{(n/2) + 1}) / 2 & n \text{ is even}\\
h_{(n+1)/2} & n \text{ is odd}
\end{cases}
\end{equation}
From the above equation, to break down the median aggregation, at least the same number of clean nodes should be perturbed. Otherwise, perturbed values will be discarded. Thus, the breakdown point of median aggregation is $1/2$. Note that we are not the first one to introduce the definition of the median on  GCNs~\cite{DBLP:conf/icassp/RuizGMR19,DBLP:journals/tsp/RuizGMR20}, but we are the first to analyze its robustness against adversarial structural perturbations from the perspective of breakdown point theory. 

\paragraph{Trimmed mean aggregation.}
Alternatively, another choice is to employ element-wise trimmed mean aggregation which eliminates extreme values during aggregation. Similar to median, the trimmed mean discards $\alpha (0 < \alpha < 0.5)$ percentage of lowest and highest values. Then it computes the mean of the remaining data. Specifically, the trimmed mean is defined as follows:
\vspace{-0.3cm}
\begin{equation}\label{eq:trimmed_mean}
a_{v} = \frac{1}{n-2\lfloor n\alpha\rfloor}\sum_{u=s}^{t}h_{u},\ s = \lfloor n\alpha\rfloor+1,\ t = n-\lfloor n\alpha\rfloor
\end{equation}
To break the trimmed mean aggregation, at least $\lfloor \alpha n\rfloor + 1$ of data points should be pushed to infinity. Otherwise, perturbed values will not be considered. Thus, the breakdown point of trimmed mean aggregation is $(\lfloor \alpha n\rfloor + 1) / n$. A larger $\alpha$ indicates a more robust aggregation. 



\section{Experiments}

\subsection{Experimental Settings}\label{sec:data}
\paragraph{Datasets.}
Following the work~\cite{DBLP:conf/kdd/ZugnerAG18,DBLP:conf/kdd/ZhuZ0019,DBLP:conf/ijcai/Wu0TDLZ19}, the experiments are conducted on four benchmark datasets, including Cora-ML~\cite{DBLP:journals/ir/McCallumNRS00}, Cora, Citeseer and Pubmed~\cite{DBLP:journals/aim/SenNBGGE08} datasets. Dataset statistics are summarized in Table~\ref{table:data}. Following the setting of previous work~\cite{DBLP:conf/kdd/ZugnerAG18},  we consider the largest connected component (LCC) of datasets. N\textsubscript{LCC} and E\textsubscript{LCC} denote the number of nodes and edges of largest connected component of each dataset. 

\begin{table}[t]
\centering
\small
\begin{tabular}{ccccc}
\hline
\textbf{Datasets} & \textbf{\#N\textsubscript{LCC}} & \textbf{\#E\textsubscript{LCC}} & \textbf{\#Features} & \textbf{\#Classes} \\
\hline
Cora & 2,485 & 5,069 & 1,433 & 7 \\
Cora-ML & 2,810 & 7,981 & 2,879 & 7 \\
Citeseer & 2,100 & 3,668 & 3,703 & 6 \\
Pubmed & 19,717 & 44,324 & 500 & 3 \\
\hline
\end{tabular}
\caption{The statistics of datasets.} \label{table:data}
\end{table}

\paragraph{Baseline models.}
To evaluate the performance and robustness, we compare our methods with four baselines: \textbf{GCN}~\cite{DBLP:conf/iclr/KipfW17}, the original GCN model which uses weighted mean to aggregate neighbors; \textbf{RGCN}~\cite{DBLP:conf/kdd/ZhuZ0019}, which models node representations as Gaussian distributions and employs a variance-based attention mechanism as the aggregation function; \textbf{Jaccard}~\cite{DBLP:conf/ijcai/Wu0TDLZ19}, that prepossesses the input graph via eliminating edges that connect nodes with Jaccard similarity of features smaller than a threshold; \textbf{SimPGCN}~\cite{simpgcn}, a similarity preserving GCN which balances the structure and feature information as well as captures pairwise node similarity to resist structural attacks. We use \textbf{TMean} and \textbf{Median} to denote GCNs that employ trimmed mean and median to aggregate neighbors. 

\paragraph{Parameter setting.}
The datasets are randomly split into training (10\%), validation (10\%), and testing (80\%) set. For each experiment, we report the average results of 5 runs. The hyper-parameters of all models are tuned based on the result on the validation set. For all models, the number of layers is set to 2 as suggested by previous works~\cite{DBLP:conf/iclr/KipfW17,DBLP:conf/iclr/VelickovicCCRLB18} and the number of hidden units is 64. We employ Adam algorithm~\cite{DBLP:journals/corr/KingmaB14} with an initial learning rate 0.01 to optimize all models. The number of training iterations is 200 with early stopping on the validation set. Following the setting of the work~\cite{DBLP:conf/ijcai/Wu0TDLZ19}, the threshold of similarity for removing dissimilar edges is set to 0. The trimmed percentage $\alpha$ of our TMean is set to 0.45 to balance the accuracy and robustness. The codes are available at \url{https://github.com/EdisonLeeeee/MedianGCN}. 

\subsection{Experimental Analysis}\label{sec:analysis}



\begin{table*}[t]
\small
\setlength{\tabcolsep}{2.0mm}
\centering
\begin{tabular}{@{}cp{1.5cm}<{\centering}p{1.5cm}<{\centering}p{1.5cm}<{\centering}p{1.5cm}<{\centering}p{1.5cm}<{\centering}p{1.8cm}<{\centering}@{}}
\toprule
\textbf{Dataset} & GCN & RGCN & Jaccard & SimPGCN & TMean & Median\\
\midrule 
\textbf{Cora} & 86.60$\pm$0.33  & 86.34$\pm$1.03 & 85.87$\pm$0.44 & 85.67$\pm$0.33 & \textbf{87.20$\pm$0.10} & 87.00$\pm$0.47\\
\textbf{Cora-ML}  & 86.00$\pm$0.26  & 85.52$\pm$0.48 & 85.15$\pm$0.26 & 85.43$\pm$0.40 & 85.67$\pm$0.12 & \textbf{86.10$\pm$0.35}\\
\textbf{Citeseer}  & 71.22$\pm$0.47  & 71.58$\pm$0.76 & 71.62$\pm$0.19 & \textbf{72.70$\pm$0.71} & 71.73$\pm$0.62 & 72.40$\pm$0.52\\
\textbf{Pubmed}  & 85.68$\pm$0.23  & 84.98$\pm$0.16 & 85.80$\pm$0.26 & \textbf{87.60$\pm$0.25} & 85.20$\pm$0.50 & 85.98$\pm$0.30\\
\bottomrule
\end{tabular}
\caption{The classification accuracy (\%) on clean datasets. The results are averaged over five runs and the best performance is boldfaced.} \label{table:acc}
\end{table*}

\begin{table*}[t]
\small
\setlength{\tabcolsep}{2.0mm}
\centering
\begin{tabular}{@{}ccp{1.4cm}<{\centering}p{1.4cm}<{\centering}p{1.4cm}<{\centering}p{1.4cm}<{\centering}p{1.4cm}<{\centering}p{1.8cm}<{\centering}@{}}
\toprule
\textbf{Dataset} & \textbf{Attack} & GCN & RGCN & Jaccard & SimPGCN & TMean & Median\\
\midrule 
\multirow{2}{*}{\textbf{Cora}}
& Direct & 2.05$\pm$0.07 & 2.79$\pm$0.10 & 3.89$\pm$0.11 & 2.12$\pm$0.09 & \textbf{4.50$\pm$0.04} & \textbf{4.59$\pm$0.07}\\
& Indirect & 9.50$\pm$0.09 & 10.11$\pm$0.06 & 10.79$\pm$0.05 & 11.27$\pm$0.15 & \textbf{11.53$\pm$0.07} & \textbf{11.52$\pm$0.06}\\
\midrule 
\multirow{2}{*}{\textbf{Cora-ML}}
& Direct & 2.95$\pm$0.01 & 3.03$\pm$0.05 & 4.32$\pm$0.03 & 2.97$\pm$0.04 & \textbf{5.51$\pm$0.02} & \textbf{5.68$\pm$0.01}\\
& Indirect & 10.08$\pm$0.06 & 10.30$\pm$0.07 & 10.13$\pm$0.10 & 11.37$\pm$0.06 & \textbf{11.55$\pm$0.04} & \textbf{11.62$\pm$0.07}\\
\midrule 
\multirow{2}{*}{\textbf{Citeseer}}
& Direct & 1.98$\pm$0.12 & 2.02$\pm$0.23 & 3.16$\pm$0.27 & 3.11$\pm$0.25 & \textbf{3.56$\pm$0.12} & \textbf{3.67$\pm$0.10}\\
& Indirect & 7.81$\pm$0.12 & 9.11$\pm$0.06 & 8.71$\pm$0.13 & \textbf{9.44}$\pm$0.17 & 9.03$\pm$0.05 & \textbf{9.43$\pm$0.06}\\
\midrule 
\multirow{2}{*}{\textbf{Pubmed}} 
& Direct & 1.14$\pm$0.02 & 1.48$\pm$0.02 & 2.13$\pm$0.03 & 2.35$\pm$0.01 & \textbf{3.80$\pm$0.04} & \textbf{3.71$\pm$0.03}\\
& Indirect & 9.12$\pm$0.01 & 10.13$\pm$0.03 & 9.37$\pm$0.04 & \textbf{12.46$\pm$0.05} & 11.79$\pm$0.03 & \textbf{11.89$\pm$0.02}\\
\bottomrule
\end{tabular}
\caption{The robustness of all models under direct and indirect attacks. The results are averaged over five runs and larger is better.} \label{table:overall}
\end{table*}

\paragraph{Accuracy.}
First, the experiments are conducted on clean datasets, which are shown in Table~\ref{table:acc}. It can be observed that our methods (i.e., TMean and Median) achieve comparable performance with baselines. TMean and Median have similar performance with original GCN, where they also achieve the best performance on Cora and Cora-ML, respectively. Even in the worst case (i.e., in Pubmed dataset), TMean and Median only slightly underperform the state-of-the-art model SimPGCN. The results demonstrate that although trimmed mean and median discard extreme values during aggregation, their accuracy is not sacrificed. 

For robustness against structural attacks, we employ NETTACK~\cite{DBLP:conf/kdd/ZugnerAG18} to perform the targeted evasion attacks on all models. In order to further explore the robustness of different models, we extend the size of target nodes set. Specifically, we randomly select 1,000 nodes from the test set and keep all the other default parameter settings in the author's implementation. As suggested in the previous works~\cite{DBLP:conf/kdd/ZhuZ0019}, the perturbation budget ranges from 1 to 5. NETTACK is employed to perform both direct and indirect attacks. The results are averaged over 5 runs. To obtain a holistic view of robustness, we employ $\sum_{q=1}^{5}q\times p_q$ as the robustness metric where $q$ is the perturbation budget and $p_q$ is the classification accuracy under the attack with perturbation budget $q$. A larger value of this metric indicating the corresponding model is more robust.

\paragraph{Overall robustness.} The results are shown in Table~\ref{table:overall}. We can observe that median aggregation outperforms all baselines under direct attack, both aggregation functions effectively preserve structural information between the nodes in the graph and their neighbors even against adversarial attacks. For indirect attack, median and trimmed mean aggregation only slightly underperforms SimPGCN on Citeseer and Pubmed. Firstly, our methods outperform RGCN in most cases, indicating attention mechanism is not the best choice for robust aggregation. Compared to the original GCN, we only change the aggregation function from the weighted mean to trimmed mean and median, demonstrating the close relationship between the robustness of the aggregation function and the whole model. Compared to indirect attacks, all models are more vulnerable to direct attacks, which is consistent with observations in the previous works~\cite{DBLP:conf/kdd/ZhuZ0019}. Generally, the Jaccard detection method achieves the best performance against direct perturbations among baselines, which verifies that perturbations are outliers in contrast to true neighbors. As expected, SimPGCN has higher robustness especially under indirect attack compared to original GCN due to preserving the node feature similarity, 
which can provide strong guidance for the model under the case of structural attacks.

\begin{figure}[ht]
  \centering
  \includegraphics[width=0.18\textwidth,trim=15 15 15 15]{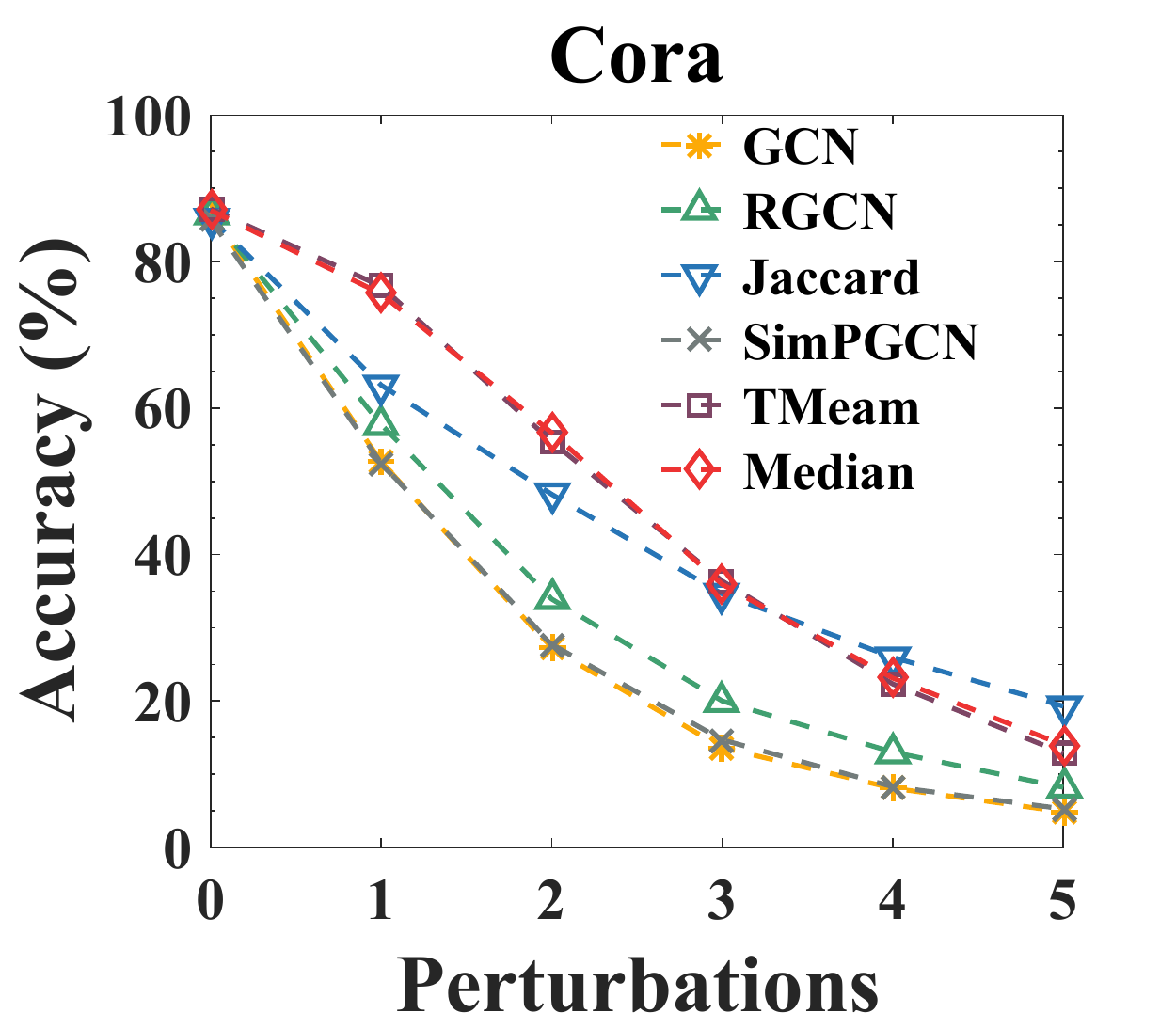}\hspace{2mm}
  \includegraphics[width=0.18\textwidth,trim=15 15 15 15]{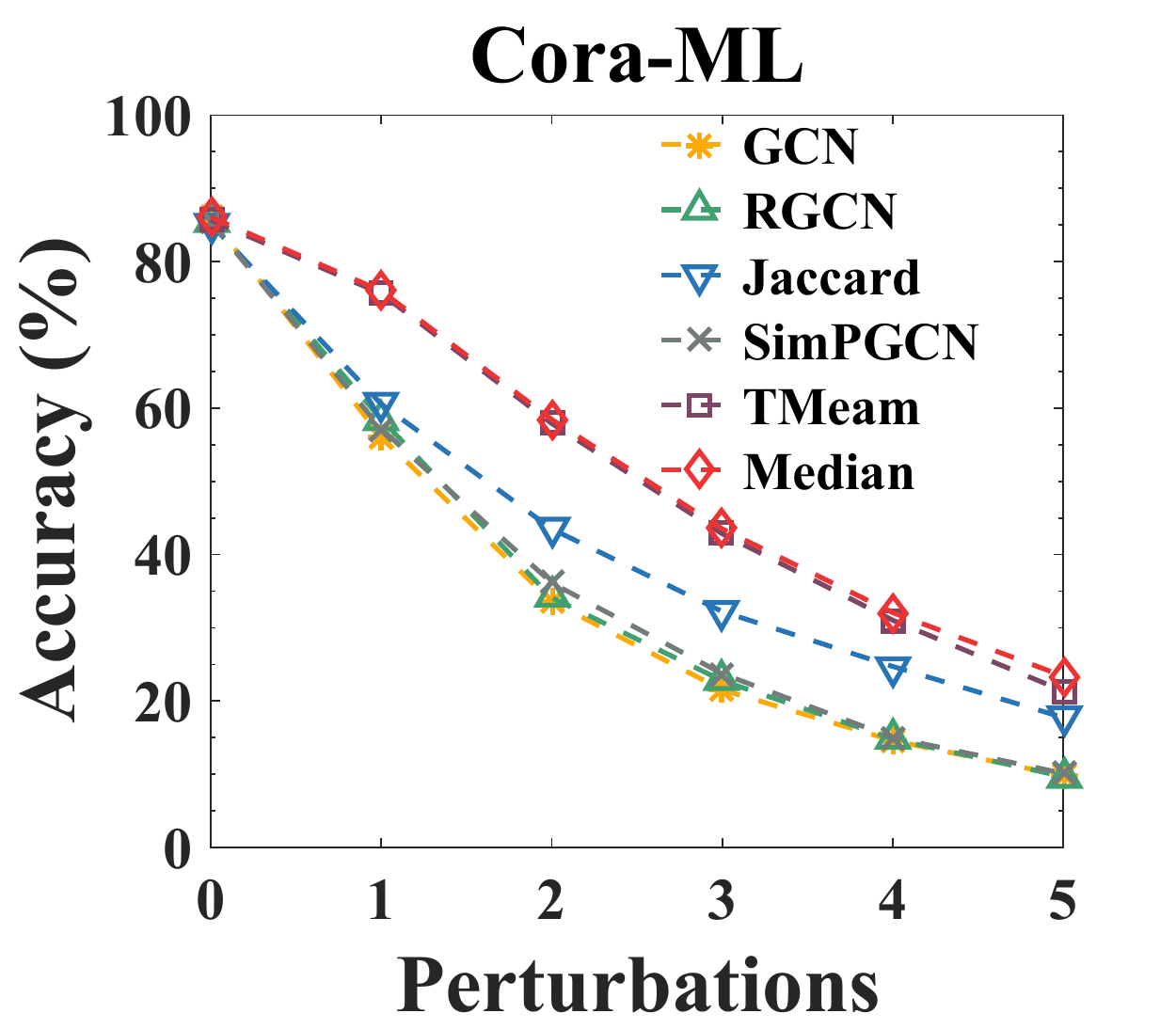}
  \caption{The robustness against direct attacks w.r.t. the number of perturbations. The results are averaged over five runs.} \label{fig:direct}
  \vspace{-0.3cm}
\end{figure}

\begin{figure}[ht]
  \centering
  \includegraphics[width=0.18\textwidth,trim=15 15 15 15]{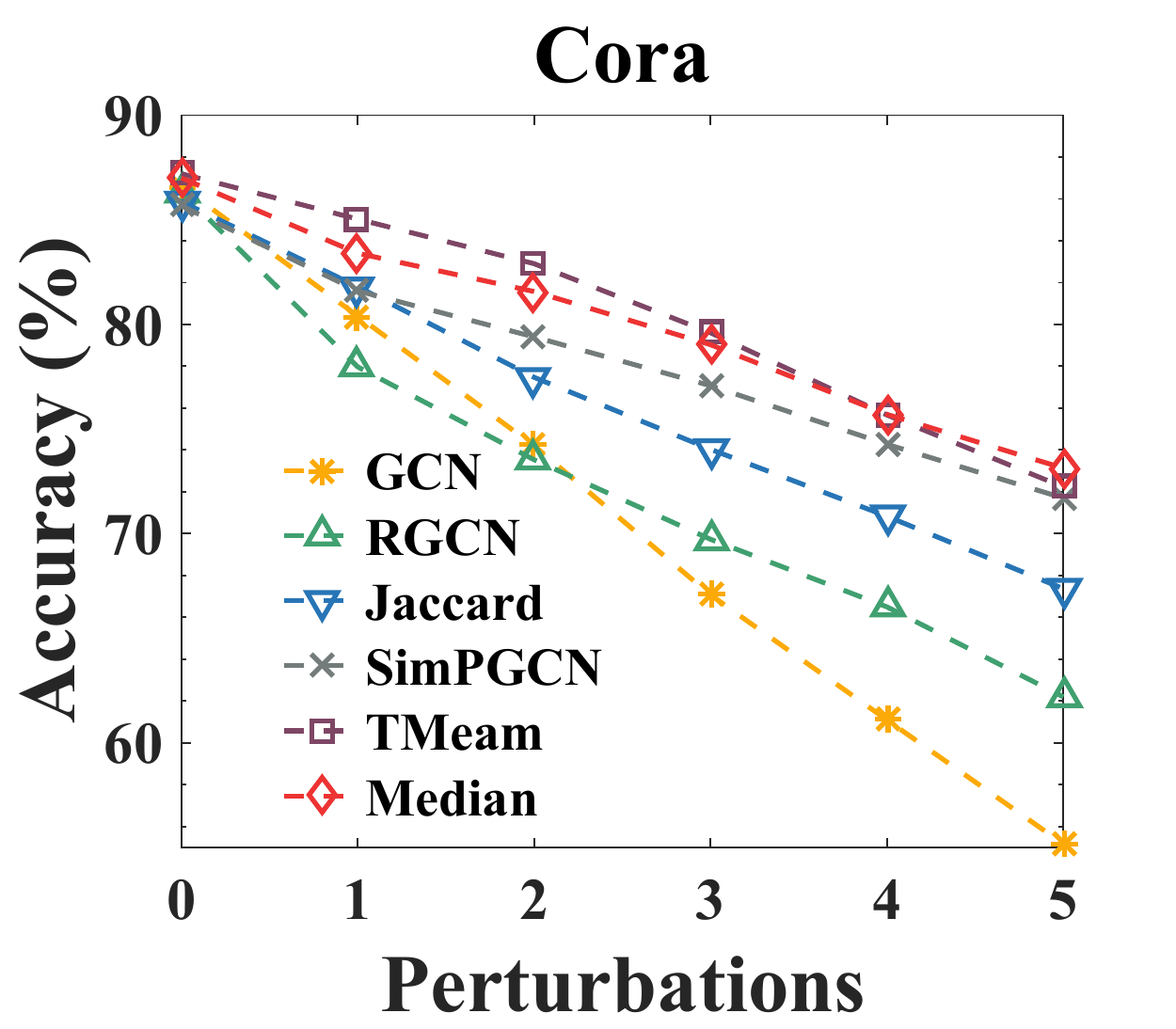}\hspace{2mm}
  \includegraphics[width=0.18\textwidth,trim=15 15 15 15]{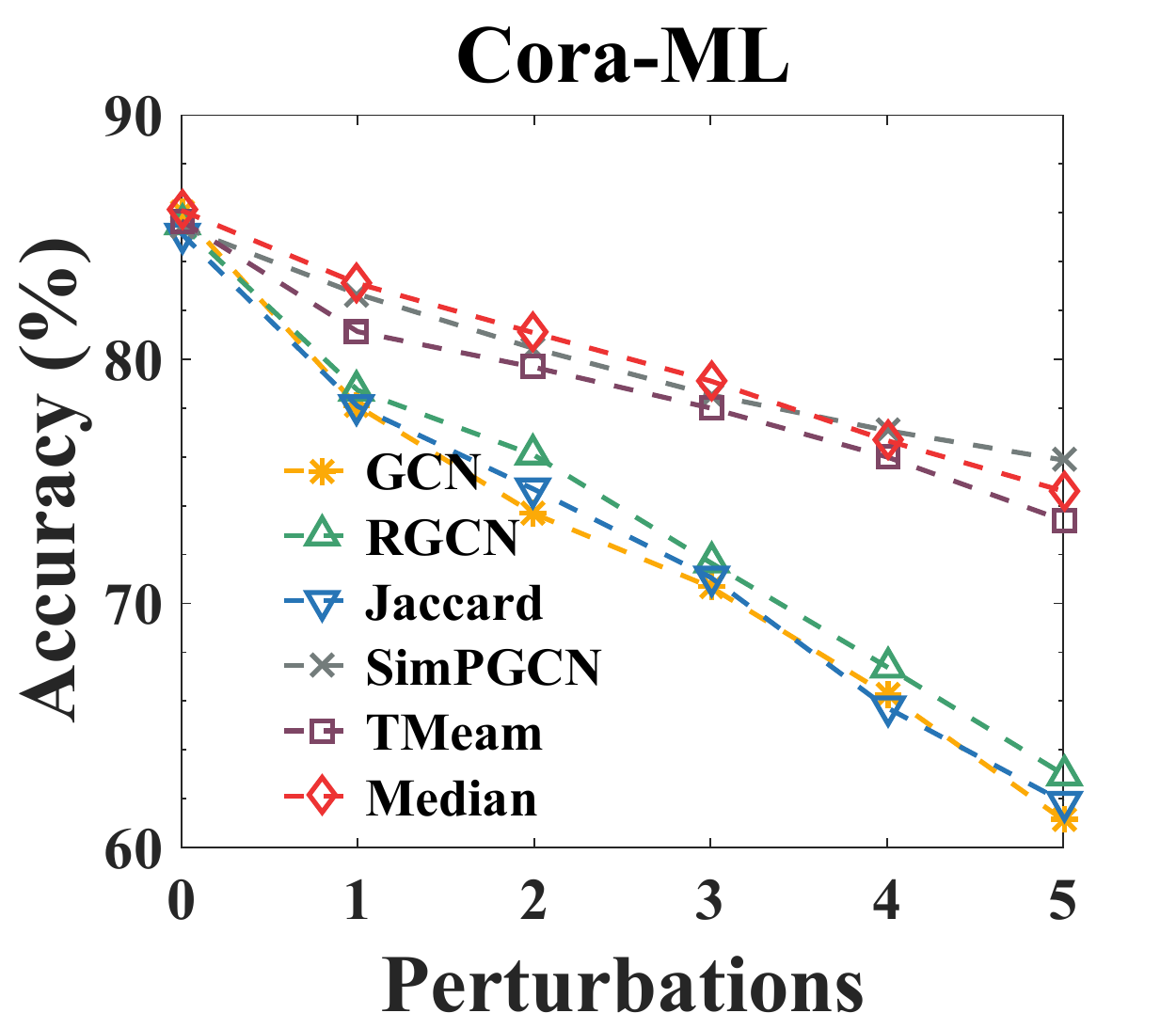}
  \caption{The robustness against indirect attacks w.r.t. the number of perturbations. The results are averaged over five runs.} \label{fig:indirect}
  \vspace{-0.3cm}
\end{figure}

\paragraph{Robustness w.r.t. the perturbation budget.}
To investigate the robustness under different number of perturbations, we plot the accuracy under direct and indirect attacks with different number of perturbations in Figure~\ref{fig:direct} and Figure~\ref{fig:indirect}.
For direct attacks, our methods generally outperform baselines. We can observe that in Figure~\ref{fig:direct}, our methods outperform baselines in all cases when the perturbation budget is 1. The reason is that the lowest degree of datasets is 2 (including self connections), thus the output is hard to be changed by a single perturbation according to that the breakdown point of median aggregation is $1/2$. When the number of perturbations is 2, the accuracy of GCN quickly drops but our methods still have high robustness. This can be attributed to the fact that the number of neighbors of most nodes is much smaller.
For indirect attacks, our methods outperform baselines in most cases, demonstrating that employing a more robust aggregation function is crucial for defending both direct and indirect attacks.

\section{Conclusion}
In this work, we aim to understand why graph convolutional networks are vulnerable to structural adversarial attacks. We highlight that one of the reasons is that the aggregation function of GCNs is non-robust, and quantitatively show that its output can be easily changed by injecting a small number of edges. According to these understandings, we propose to use more robust aggregation functions such as trimmed mean and median. Extensive experiments demonstrate that such simple but effective methods achieve state-of-the-art robustness without sacrificing accuracy. 
For future work, we will extend our study to analyze the non-targeted attacks on GCNs and find whether there is possible to develop models that are robust against both the targeted and non-targeted attacks.

\section*{Acknowledgements}
The research is supported by the Key-Area Research and Development Program of Guangdong Province (No. 2020B010165003), the National Natural Science Foundation of China (No. U1711267),  the Guangdong Basic and Applied Basic Research Foundation (No. 2020A1515010831), and the Program for Guangdong Introducing Innovative and Entrepreneurial Teams (No. 2017ZT07X355).
\bibliographystyle{named}
\bibliography{ijcai21}

\begin{thebibliography}{}

\bibitem[\protect\citeauthoryear{Bojchevski and
  G{\"{u}}nnemann}{2019a}]{DBLP:conf/icml/BojchevskiG19}
Aleksandar Bojchevski and Stephan G{\"{u}}nnemann.
\newblock Adversarial attacks on node embeddings via graph poisoning.
\newblock In {\em ICML}, pages 695--704, 2019.

\bibitem[\protect\citeauthoryear{Bojchevski and
  G{\"{u}}nnemann}{2019b}]{DBLP:conf/nips/BojchevskiG19}
Aleksandar Bojchevski and Stephan G{\"{u}}nnemann.
\newblock Certifiable robustness to graph perturbations.
\newblock In {\em NeurIPS}, pages 8317--8328, 2019.

\bibitem[\protect\citeauthoryear{Chen \bgroup \em et al.\egroup
  }{2020}]{chen2020survey}
Liang Chen, Jintang Li, Jiaying Peng, Tao Xie, Zengxu Cao, Kun Xu, Xiangnan He,
  and Zibin Zheng.
\newblock A survey of adversarial learning on graph.
\newblock {\em arXiv preprint arXiv:2003.05730}, 2020.

\bibitem[\protect\citeauthoryear{Dai \bgroup \em et al.\egroup
  }{2018}]{DBLP:conf/icml/DaiLTHWZS18}
Hanjun Dai, Hui Li, Tian Tian, Xin Huang, Lin Wang, Jun Zhu, and Le~Song.
\newblock Adversarial attack on graph structured data.
\newblock In {\em ICML}, pages 1123--1132, 2018.

\bibitem[\protect\citeauthoryear{Donoho \bgroup \em et al.\egroup
  }{1992}]{donoho1992breakdown}
David~L Donoho, Miriam Gasko, et~al.
\newblock Breakdown properties of location estimates based on halfspace depth
  and projected outlyingness.
\newblock {\em The Annals of Statistics}, 20(4):1803--1827, 1992.

\bibitem[\protect\citeauthoryear{Entezari \bgroup \em et al.\egroup
  }{2020}]{DBLP:conf/wsdm/EntezariADP20}
Negin Entezari, Saba~A. Al{-}Sayouri, Amirali Darvishzadeh, and Evangelos~E.
  Papalexakis.
\newblock All you need is low (rank): Defending against adversarial attacks on
  graphs.
\newblock In {\em WSDM}, pages 169--177, 2020.

\bibitem[\protect\citeauthoryear{Fan \bgroup \em et al.\egroup
  }{2019}]{DBLP:conf/www/Fan0LHZTY19}
Wenqi Fan, Yao Ma, Qing Li, Yuan He, Yihong~Eric Zhao, Jiliang Tang, and Dawei
  Yin.
\newblock Graph neural networks for social recommendation.
\newblock In {\em WWW}, pages 417--426, 2019.

\bibitem[\protect\citeauthoryear{Geisler \bgroup \em et al.\egroup
  }{2020}]{GeislerZG20}
Simon Geisler, Daniel Z{\"{u}}gner, and Stephan G{\"{u}}nnemann.
\newblock Reliable graph neural networks via robust aggregation.
\newblock In {\em NeurIPS}, 2020.

\bibitem[\protect\citeauthoryear{Huber and Ronchetti}{2009}]{Huber:1254106}
Peter~J Huber and Elvezio~M Ronchetti.
\newblock {\em Robust statistics; 2nd ed.}
\newblock Wiley, Hoboken, NJ, 2009.

\bibitem[\protect\citeauthoryear{Jin \bgroup \em et al.\egroup
  }{2020}]{DBLP:journals/corr/abs-2003-00653}
Wei Jin, Yaxin Li, Han Xu, Yiqi Wang, and Jiliang Tang.
\newblock Adversarial attacks and defenses on graphs: {A} review and empirical
  study.
\newblock {\em CoRR}, abs/2003.00653, 2020.

\bibitem[\protect\citeauthoryear{Jin \bgroup \em et al.\egroup
  }{2021}]{simpgcn}
Wei Jin, Tyler Derr, Yiqi Wang, Yao Ma, Zitao Liu, and Jiliang Tang.
\newblock Node similarity preserving graph convolutional networks.
\newblock In {\em WSDM}, 2021.

\bibitem[\protect\citeauthoryear{Kingma and
  Ba}{2015}]{DBLP:journals/corr/KingmaB14}
Diederik~P. Kingma and Jimmy Ba.
\newblock Adam: {A} method for stochastic optimization.
\newblock In {\em ICLR}, 2015.

\bibitem[\protect\citeauthoryear{Kipf and
  Welling}{2017}]{DBLP:conf/iclr/KipfW17}
Thomas~N. Kipf and Max Welling.
\newblock Semi-supervised classification with graph convolutional networks.
\newblock In {\em ICLR}, 2017.

\bibitem[\protect\citeauthoryear{Liu \bgroup \em et al.\egroup
  }{2020}]{DBLP:conf/ecai/LiuPCZ20}
Yang Liu, Jiaying Peng, Liang Chen, and Zibin Zheng.
\newblock Abstract interpretation based robustness certification for graph
  convolutional networks.
\newblock In {\em {ECAI}}, volume 325, pages 1309--1315. {IOS} Press, 2020.

\bibitem[\protect\citeauthoryear{McCallum \bgroup \em et al.\egroup
  }{2000}]{DBLP:journals/ir/McCallumNRS00}
Andrew McCallum, Kamal Nigam, Jason Rennie, and Kristie Seymore.
\newblock Automating the construction of internet portals with machine
  learning.
\newblock {\em Inf. Retr.}, 3(2):127--163, 2000.

\bibitem[\protect\citeauthoryear{Ruiz \bgroup \em et al.\egroup
  }{2019}]{DBLP:conf/icassp/RuizGMR19}
Luana Ruiz, Fernando Gama, Antonio~G. Marques, and Alejandro Ribeiro.
\newblock Median activation functions for graph neural networks.
\newblock In {\em ICASSP}, pages 7440--7444, 2019.

\bibitem[\protect\citeauthoryear{Ruiz \bgroup \em et al.\egroup
  }{2020}]{DBLP:journals/tsp/RuizGMR20}
Luana Ruiz, Fernando Gama, Antonio~Garcia Marques, and Alejandro Ribeiro.
\newblock Invariance-preserving localized activation functions for graph neural
  networks.
\newblock {\em {IEEE} Trans. Signal Process.}, 68:127--141, 2020.

\bibitem[\protect\citeauthoryear{Sen \bgroup \em et al.\egroup
  }{2008}]{DBLP:journals/aim/SenNBGGE08}
Prithviraj Sen, Galileo Namata, Mustafa Bilgic, Lise Getoor, Brian Gallagher,
  and Tina Eliassi{-}Rad.
\newblock Collective classification in network data.
\newblock {\em {AI} Magazine}, 29(3):93--106, 2008.

\bibitem[\protect\citeauthoryear{Tang \bgroup \em et al.\egroup
  }{2020}]{DBLP:conf/wsdm/TangLSYMW20}
Xianfeng Tang, Yandong Li, Yiwei Sun, Huaxiu Yao, Prasenjit Mitra, and Suhang
  Wang.
\newblock Transferring robustness for graph neural network against poisoning
  attacks.
\newblock In {\em WSDM}, pages 600--608, 2020.

\bibitem[\protect\citeauthoryear{Velickovic \bgroup \em et al.\egroup
  }{2018}]{DBLP:conf/iclr/VelickovicCCRLB18}
Petar Velickovic, Guillem Cucurull, Arantxa Casanova, Adriana Romero, Pietro
  Li{\`{o}}, and Yoshua Bengio.
\newblock Graph attention networks.
\newblock In {\em ICLR}, 2018.

\bibitem[\protect\citeauthoryear{Wang \bgroup \em et al.\egroup
  }{2019}]{DBLP:conf/icdm/WangQL0JWFYZY19}
Daixin Wang, Yuan Qi, Jianbin Lin, Peng Cui, Quanhui Jia, Zhen Wang, Yanming
  Fang, Quan Yu, Jun Zhou, and Shuang Yang.
\newblock A semi-supervised graph attentive network for financial fraud
  detection.
\newblock In {\em ICDM}, pages 598--607, 2019.

\bibitem[\protect\citeauthoryear{Wu \bgroup \em et al.\egroup
  }{2019}]{DBLP:conf/ijcai/Wu0TDLZ19}
Huijun Wu, Chen Wang, Yuriy Tyshetskiy, Andrew Docherty, Kai Lu, and Liming
  Zhu.
\newblock Adversarial examples for graph data: Deep insights into attack and
  defense.
\newblock In {\em IJCAI}, pages 4816--4823, 2019.

\bibitem[\protect\citeauthoryear{Xie \bgroup \em et al.\egroup
  }{2020}]{xie2020gnns}
Yiqing Xie, Sha Li, Carl Yang, Raymond Chi-Wing Wong, and Jiawei Han.
\newblock When do gnns work: Understanding and improving neighborhood
  aggregation.
\newblock In {\em IJCAI}, volume 2020, 2020.

\bibitem[\protect\citeauthoryear{Xu \bgroup \em et al.\egroup
  }{2019a}]{DBLP:conf/ijcai/XuC0CWHL19}
Kaidi Xu, Hongge Chen, Sijia Liu, Pin{-}Yu Chen, Tsui{-}Wei Weng, Mingyi Hong,
  and Xue Lin.
\newblock Topology attack and defense for graph neural networks: An
  optimization perspective.
\newblock In {\em IJCAI}, pages 3961--3967, 2019.

\bibitem[\protect\citeauthoryear{Xu \bgroup \em et al.\egroup
  }{2019b}]{DBLP:conf/iclr/XuHLJ19}
Keyulu Xu, Weihua Hu, Jure Leskovec, and Stefanie Jegelka.
\newblock How powerful are graph neural networks?
\newblock In {\em ICLR}, 2019.

\bibitem[\protect\citeauthoryear{Zhu \bgroup \em et al.\egroup
  }{2019}]{DBLP:conf/kdd/ZhuZ0019}
Dingyuan Zhu, Ziwei Zhang, Peng Cui, and Wenwu Zhu.
\newblock Robust graph convolutional networks against adversarial attacks.
\newblock In {\em KDD}, pages 1399--1407, 2019.

\bibitem[\protect\citeauthoryear{Z{\"{u}}gner and
  G{\"{u}}nnemann}{2019a}]{DBLP:conf/iclr/ZugnerG19}
Daniel Z{\"{u}}gner and Stephan G{\"{u}}nnemann.
\newblock Adversarial attacks on graph neural networks via meta learning.
\newblock In {\em ICLR}, 2019.

\bibitem[\protect\citeauthoryear{Z{\"{u}}gner and
  G{\"{u}}nnemann}{2019b}]{DBLP:conf/kdd/ZugnerG19}
Daniel Z{\"{u}}gner and Stephan G{\"{u}}nnemann.
\newblock Certifiable robustness and robust training for graph convolutional
  networks.
\newblock In {\em KDD}, pages 246--256, 2019.

\bibitem[\protect\citeauthoryear{Z{\"{u}}gner \bgroup \em et al.\egroup
  }{2018}]{DBLP:conf/kdd/ZugnerAG18}
Daniel Z{\"{u}}gner, Amir Akbarnejad, and Stephan G{\"{u}}nnemann.
\newblock Adversarial attacks on neural networks for graph data.
\newblock In {\em KDD}, pages 2847--2856, 2018.

\end{thebibliography}

\end{document}